\pdfoutput=1

\documentclass[11pt]{article}
\usepackage{amsmath}
\usepackage{color}
\usepackage{amssymb}
\usepackage{tabularx}
\usepackage{multirow}
\usepackage{graphicx}
\usepackage{stfloats}
\usepackage{booktabs} 
\usepackage{bbding}
\usepackage{multicol}
\usepackage[dvipsnames]{xcolor}
\usepackage{threeparttable} 
\usepackage{url}
\usepackage{float}
\usepackage{enumitem}

\usepackage[]{acl}

\usepackage{times}
\usepackage{latexsym}

\usepackage[T1]{fontenc}

\usepackage[utf8]{inputenc}

\usepackage{microtype}

\title{CGoDial: A Large-Scale Benchmark for Chinese Goal-oriented Dialog Evaluation} 

\author{
Yinpei Dai, Wanwei He, Bowen Li, Yuchuan Wu, Zheng Cao, \\
\textbf{Zhongqi An, Jian Sun, Yongbin Li\thanks{$^*$Corresponding author}}\\
    \text{Alibaba Group, Beijing, China}\\
    \texttt{\{yinpei.dyp,hewanwei.hww,shengxiu.wyc,zhengzhi.cz,} \\
     \texttt{zhongqi.azq,jian.sun,shuide.lyb\}@alibaba-inc.com}
}

\begin{document}
\maketitle
\begin{abstract}
Practical dialog systems need to deal with various knowledge sources, noisy user expressions, and the shortage of annotated data. To better solve the above problems, we propose \textbf{CGoDial}\footnote{Dataset available at \texttt{https://github.com/Alibaba\\Research/DAMO-ConvAI/cgodial}.}, a new challenging and comprehensive \textbf{\underline{C}}hinese benchmark for multi-domain  \textbf{\underline{Go}}al-oriented \textbf{\underline{Dial}}og evaluation. 
It contains 96,763 dialog sessions, and 574,949 dialog turns totally, covering three datasets with different knowledge sources: 1) a slot-based dialog (SBD) dataset with table-formed knowledge, 2) a flow-based dialog (FBD) dataset with tree-formed knowledge, and a retrieval-based dialog (RBD) dataset with candidate-formed knowledge. To bridge the gap between academic benchmarks and spoken dialog scenarios, we either collect data from real conversations or add spoken features to existing datasets via crowd-sourcing. 
The proposed experimental settings include the combinations of training with either the entire training set or a few-shot training set, and testing with either the standard test set or a hard test subset, which can assess model capabilities in terms of general prediction, fast adaptability and reliable robustness. 
\end{abstract}

\section{Introduction}
Goal-oriented dialog systems converse with users naturally, helping them fulfill specific goals such as restaurant booking, hotel reservation, and flight search. 
Many popular datasets have been introduced to facilitate the dialog research, ranging from simple single-domain \cite{wen-etal-2017-network} to more difficult multi-domain dialogs \cite{budzianowski-etal-2018-multiwoz}. 
Similar to the widely studied GLUE \cite{wang2018glue} and SuperGLUE \cite{wang2019superglue} for general natural language understanding, large-scale benchmarks have been specifically designed for dialog systems. For instance, 
\citet{mehri2020dialoglue} proposed the DialoGLUE, a collection of seven popular English dialog datasets, to assess the  natural language understanding ability in a few-shot setting. More recently, 
\citet{peng-etal-2021-raddle} introduced the RADDLE, another large-scale benchmark that includes a diagnostic checklist facilitating detailed robustness analysis for pre-trained language models \cite{radford2019language, peng-etal-2021-soloist}.

However, compared to the abundance of English dialog benchmarks, the resource of Chinese dialog datasets in goal-oriented dialog remains quite limited, let alone the large-scale benchmark suitable for the dialog evaluation of Chinese pre-trained language models. 
Moreover, current Chinese dialog datasets, such as CrossWOZ \cite{zhu-etal-2020-crosswoz} and RiSAWOZ \cite{quan-etal-2020-risawoz}, mainly focus on the schema-guided dialog as in the classical MultiWOZ dataset \cite{budzianowski-etal-2018-multiwoz}, where conversations are grounded on a table-formed ontology. This kind of ontology usually  defines a set of possible slot-value pairs to be recognized during the dialog for database query \cite{young2013pomdp}.
In real-world applications, however, more types of goal-oriented dialogs are involved. 
For instance, in \citet{xie2022converse} the conversational process is grounded on a pre-defined tree-formed ontology, where the whole process is constrained in a dialog flow.
Also, in some scenarios, the conversation has been simplified into a multi-turn response selection problem so that the dialog system is able to transfer to new domains quickly and reliably \citep{henderson-etal-2019-repository, henderson-etal-2020-convert, dai-etal-2020-learning}.

These observations motivate us to construct CGoDial, a new large-scale Chinese benchmark for goal-oriented dialog,  targeting the evaluation of model adaptability and robustness in real situations.
Concretely, we consider three types of goal-oriented dialogs with different  knowledge database sources.
The first is the slot-based dialog (SBD), where a system often asks a user for required attributes to constrain the search in a table-formed database and then provides entities and extra information to the user. To build SBD,  we adapt the existing Chinese dataset RiSAWOZ \cite{quan-etal-2020-risawoz} by supplementing it using crowd-sourcing for more difficult variations such as noisy expression augmentation, external knowledge utilization, and out-of-scope management. 44,800 new dialogs spanning 12 domains are collected for SBD.
The second is the flow-based dialog (FBD), where the system guides the user to fulfill specific tasks based on tree-structured dialog flow. This flow is a new form of knowledge database that defines causal constraints of conversations. For example, when a user comes to withdraw money in a housing insurance consulting scenario, the system must first determine the user's identity before moving to the next step. Therefore,  
compared with SBD, FBD has a strict order to request user information
We collect new dialogs between human customers and an online customer service dialog agent from real businesses, which to the best of our knowledge, is the first Chinese FBD dataset. After data desensitization, we have 6,786 dialogs and 45,601 turns spanning four different domains in the end.
The third is the retrieval-based dialog (RBD), like bAbI-dialog \cite{bordes2016learning}, where the dialog system learns to select the correct response from a candidate response set. We build RBD by adapting the existing  ECD \cite{zhang-etal-2018-modeling} dataset and adding more noisy user expressions like ASR errors to raise the overall difficulty. We choose 56,000 non-ambiguous complex dialog examples from the original 
dataset.

To fully assess dialog model capabilities in general prediction, fast adaptability, and reliable robustness. For all the above types of dialog datasets, we propose four different experimental settings, including the combinations of training with either the full training set or a few-shot training set, and testing with either the standard test set or a hard test subset.
Extensive experiments have been conducted on CGoDial with various Chinese pre-trained language model baselines for dialog modeling, such as Chinese-T5 \cite{zhao-etal-2019-uer} and CDial-GPT \cite{wang2020large}. To further facilitate the research of dialog pre-training, we also release a new UniLM-based dialog model pre-trained on large-scale human-to-human dialog corpora, which achieves the best results on all tasks. 

Therefore, the contributions of this paper are three-fold:

\begin{itemize}
    \item A challenging and comprehensive Chinese dialogue benchmark consisting of three different types of goal-oriented dialogs.
    \item  Standardized evaluation measures that facilitate the study of robustness and adaptability.
    \item Competitive baselines across different dialog tasks. Both the datasets and codes will be open-sourced to push forward the research in goal-oriented dialog. 
\end{itemize}


\section{Related Work}

\subsection{Goal-oriented Dialog Benchmarks.}
The introduction of benchmarks brings forward the research of goal-oriented dialog. Several recent tendencies have penetrated the development:
\paragraph{From single-domain to multi-domain.} The earliest dataset such as bAbI-dialog \cite{bordes2016learning} and WOZ \cite{wen-etal-2017-network} focus on the single-domain dialog. In the following, extended-bAbI \cite{dai-etal-2020-learning},  MultiWOZ \cite{budzianowski-etal-2018-multiwoz} and Schema-Guided Dialog \cite{mosig2020star} are proposed to solve more difficult tasks in multi-domain dialog. 
\paragraph{From slot-based to flow-based.} The classical MultiWOZ dataset is grounded on the table-formed ontology, so the dialog understanding problem can be put forward as a multi-turn slot-filling task. However, in many real applications, there are fixed order constraints to collect the slot information from users, STAR \cite{mosig2020star} and ABCD \cite{chen-etal-2021-action} are proposed recently to cover this property by exerting causal constraints on the dialog flows.
\paragraph{From simplified tasks to real scenarios.}  Most previous dialog benchmarks \cite{rojas2018deep, dai2018tracking, rastogi2020towards, quan-etal-2020-risawoz, zhu-etal-2020-crosswoz, zhang2022layout} focus on text-in text-out dialogs but neglect spoken characteristics in real problems. RADDLE \cite{peng-etal-2021-raddle} is the first to consider evaluating model robustness by adding various ASR noises to the original MultiWOZ but is not publicly available now.  \citet{SADSTC} extends the same idea to propose a speech-aware dialog systems technology challenge. 
NSD \cite{wu-etal-2021-novel} aims to discover unknown or out-of-domain slot types for dialogue systems. EmoWOZ \cite{feng2021emowoz} recognizes the critical role of emotion and provides a new emotion-annotated corpus of goal-oriented dialogs based on MultiWOZ. SSTOD \cite{zhang2022slot} proposes a novel sub-slot filling task that is crucial in string-formed information collection like phone numbers and people names.

\paragraph{From single-modal to multi-modal.} One type of multi-modal dialog is to build a system that can help users search for target objects via generating textual responses and object pictures.
Common datasets are MMD \cite{saha2018towards}, MMConv \cite{liao2021mmconv} and JDDC 2.0 \cite{zhao2021jddc}, another type is to deal with immersive and situated multi-modal scenarios, such as SIMMC \cite{crook2019simmc}, SIMMC2.0 \cite{kottur-etal-2021-simmc} and TEACH \cite{padmakumar2021teach}, where the visual input of the agent's surrounding environment needs to be used for conversations; and 5) From monolingual to multi-lingual, such as BiTOD \cite{lin2021bitod}, GlobalWOZ \cite{ding2021globalwoz}, Multi$^2$WOZ \cite{hung2022multi2woz} and  AllWOZ \cite{zuo2021allwoz}. 

Compared with previous work, our CGoDial focuses on single-modal and is the first Chinese benchmark that covers flow-based dialog and considers real scenarios.

\subsection{Dialog Data Collection}
Goal-oriented dialog construction can be broadly divided into three categories according to the data collection scheme.

\paragraph{Machine-to-machine (M2M) scheme}  creates data via dialog simulation \cite{bordes2016learning, rastogi2020towards, kottur-etal-2021-simmc}, given manually designed utterance templates and dialogue process. Crowd-sourcing is then used to paraphrase the dialog with more varied language expressions.  This approach can collect very large-scale data at low cost \cite{shah2018building} but often lacks noisy conditions and flexible processes that appear in human conversations \cite{black2011spoken}.
\paragraph{Human-to-machine (H2M) scheme} collects dialogs based on an existing dialog system. Standard datasets include the Let’s Go Bus Information System \cite{raux2005let} and the second and third Dialog State Tracking Challenges \cite{henderson2014second,henderson2014third}. This method is closest to the real applications but requires a well-deployed dialog agent  prepared ahead.
\paragraph{Human-to-human (H2H) scheme} is also a cheap way to collect dialog data as there are many available dialog corpora on the internet like Twitter \cite{ritter2010unsupervised} and Reddit \cite{henderson2019repository}. However, building annotated dialogs that meet the requirements of specific tasks relies on the costly Wizard-of-Oz framework. CrossWOZ \cite{zhu-etal-2020-crosswoz} and MultiWOZ \cite{budzianowski-etal-2018-multiwoz} lie in this line. 

In our CGoDial benchmark,  slot-based dialog and retrieval-based dialog are collected by H2H, and flow-based dialog is collected by H2M.

\section{CGoDial Benchmark}
\begin{figure*}
    \centering
    \includegraphics[width=1.0\textwidth]{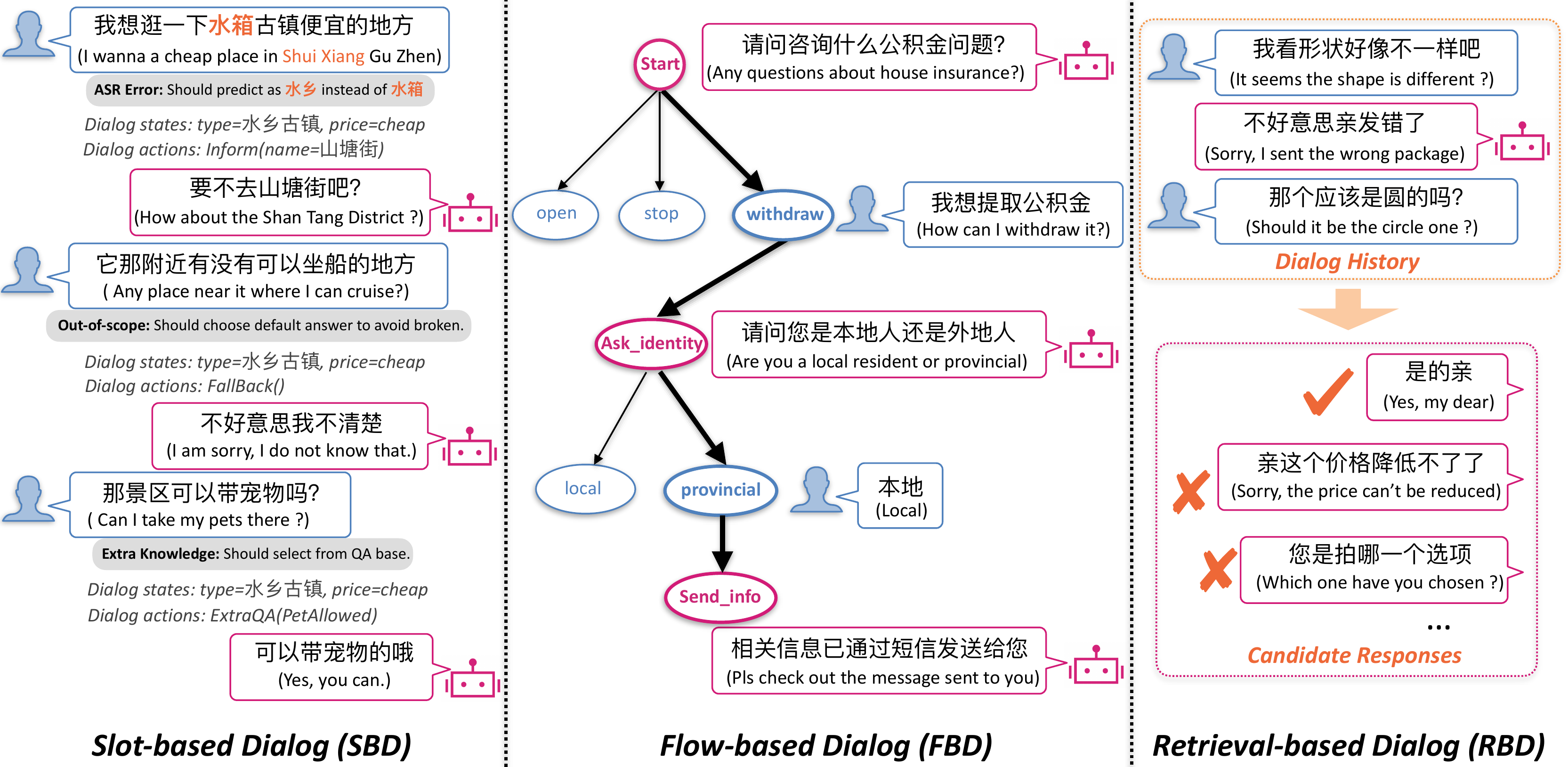}
    \caption{Three dialog examples of SBD, FBD and RBD in CGoDial, respectively.}
    \label{fig:dataset}
\end{figure*}

In this section, we elaborate on the three datasets in our  CGoDial benchmark, in terms of task definition and dataset construction. To give an overall impression, Figure \ref{fig:dataset} illustrate typical examples for each dataset and Table \ref{tab: dataset} summarizes detailed statistics. 
Appendix A.1 illustrates more dataset distributions of dialog length and turn length.

\begin{table*}[t]
\scalebox{0.9}{
\begin{tabular}{cccccc}
\toprule
Dataset              &         & Train   & Valid  & Test-std   & Test-hard \\ \hline
\multirow{2}{*}{Slot-Based Dialog (SBD)} & \#dialog & 40,000  & 2,400  & 2,400  & 600   \\
                     & \#turn   & 403,740 & 32,464 & 37,144 & 9,286 \\
\multirow{2}{*}{Flow-Based Dialog (FBD)} & \#dialog & 3,950   & 1,450  & 1,386  & 985   \\
                     & \#turn   & 15,558  & 5,167  & 5,624  & 3,629 \\
\multirow{2}{*}{Retrieval-Based Dialog (RBD)} & \#dialog & 39,589  & 4,627   & 961    & 701   \\
                     & \#turn   & 50,000  & 5,000  & 1,000  & 719   \\ \bottomrule
\end{tabular}
}
\centering
\caption{Statistics of CGoDial. We maintain two test sets for evaluation. One is the standard test set (denoted as test-std), and another is a robust test set that is a hard subset chosen from test-std (denoted as test-hard).}
\label{tab: dataset}
\end{table*}

\subsection{Slot-based Dialog Dataset}
The slot-based dialog (SBD) is tasked to search and provide entities from a  table-formed database that meet the requirement of a user through natural interactions, such as reserving a restaurant for the user. The database specifies an ontology about the semantic scope of the system can process, which is represented as a set of slot-value pairs to maintain a dialog state at each turn. For example, in the classical MultiWOZ dataset \cite{budzianowski-etal-2018-multiwoz}, if a user says \textit{``I would like a cheap hotel in the west."}, the system needs to extract the dialog state as \textit{`pricerange=cheap, area=west'}. Then the system uses the dialog state to query the database and decides dialog actions based on the searching results. For example, the system can decide to request more unfilled slots like \textit{``what type of food do you like"} to constrain the search if there are too many returned entities. After the entity is fixed, the user may ask for more  information about the entity like \textit{``what is the phone number and address of the nados restaurant?"}, and the system needs to provide the information of the requested slots  \textit{`phone\_number, address'}.


\subsubsection{Task Definition} 
To fully evaluate the model capability, we include tasks for dialog understanding, policy, and generation. For the dialog understanding, we choose the dialog state tracking (DST) task \cite{henderson2014second}, which tracks the slot-values conveyed by the user during the dialog. We use \texttt{JGA} to denote the joint goal accuracy as in  \citet{heck-etal-2020-trippy} for DST, which is the proportion of turns that all slot-value pairs have been correctly predicted.
For the dialog policy, since there is the one-to-many property \cite{zhang2020task} in policy planning and we do not have user simulators   \cite{ultes-etal-2017-pydial} to calculate the online task success rate,  we adopt a similar metric in \citet{budzianowski-etal-2018-multiwoz} and use \texttt{Succ} to denote turn accuracy of dialogs where all user requested slots\footnote{Except regular requested slots in RiSAWOZ, external QA pairs and OOS utterances are also treated as specially requested
slots to predict together. Details are given in the next section.} have been correctly answered and the searched entity follows the oracle dialog state. 
For the dialog generation, we follow the common practice to compute the average BLEU scores \cite{papineni-etal-2002-bleu} for all turns. Specifically, we choose BLEU-4 in SBD.  
Inspire by \citet{mehri-etal-2019-structured}, we also calculate a combined score \texttt{Comb} to measure the overall performance for all tasks, which is the geometric mean of above metrics: \texttt{Comb}=$^{3}\sqrt{\texttt{JGA}\times \texttt{Succ} \times \texttt{BLEU}}$. Different from the original combined score that adds all scores linearly, the geometric mean should be more reasonable in the aspect of dimensional calculation. 

\subsubsection{Dataset Construction}
We build our SBD dataset based on the existing Chinese dataset RiSAWOZ \cite{quan-etal-2020-risawoz}. Since RiSAWOZ is purely text-in-text-out and has limited language variation, it can not reflect the difficulty in realistic spoken dialog system applications.  We add three important common features in real scenarios to fill this gap in the current Chinese dialog benchmarks.

\paragraph{Feature 1: External knowledge.}  Users often ask for some new knowledge that is not covered in the current ontology. Inspired by the work of \cite{kim-etal-2020-beyond}, we expand the coverage of dialog systems by incorporating external unstructured knowledge sources to tackle users' unseen requests. In our SBD, the external knowledge is represented as question-answer (QA) pairs that can be utilized during the conversation. The system must decide whether the current turn should select a proper answer from the external resources or generate the answer by itself. To construct this new part of dialog data,  we first ask the crowd-sourcing people to make up new relevant QA pairs given the RiSAWOZ dialogs and the ontology. After manually selecting 150 basic QA pairs, such as \textit{``Can I take my pets?"} and  \textit{``Are there any artistic shows in the scenic spot?"}, we ask the crowd-sourcing people to paraphrase each QA pair into nine similar pairs, and acquire a total of 1,500 QA pairs. Then we insert QA pairs into the original dialog randomly as the external knowledge. Specifically, each dialog is inserted at least one QA pair with a probability of 0.5. We also hold 200 QA pairs out of the training or validation sets, and only add them into the test set for generalization evaluation. 
We treat the question selection problem as specially requested slots detection problem like \textit{`IsPetAllowed'}, and count the accuracy as a part of \texttt{Succ}.

\paragraph{Feature 2: Out-of-scope (OOS) utterances.} One of the most common problems in practical dialog systems is that users can talk about something beyond the semantic scope that the system can process (e.g., meaningless sentences like \textit{`Uh huh... well it should have...'} and irrelevant questions \textit{`where can I cruise'}). Thus, practical dialog systems require robust detection of OOS situations to avoid conversational breakdowns and properly handle unseen user behaviors.
To simulate the feature, for each dialog in RiSAWOZ, we ask the human annotators to insert plausible OOS turns based on the given context and all external QA pairs. Each dialog is inserted at most two OOS utterances with the probability of 0.6. The corresponding default answer for OOS is \textit{``I am sorry, I do not know that''}. Like QA prediction, we treat the OOS detection as a special requested slot  \textit{`OOS'} detection, and count the accuracy as a part of \texttt{Succ}.

\paragraph{Feature 3: Spoken noise.} 
To mimic the spoken language phenomena in real applications, we add speech errors by crowd-sourcing. More concretely, we ask the people to read out all utterances four times in modified RiSAWOZ, allowing them to vary the expression subtly under the same core semantics. For example, the utterance \textit{`I want a cheap hotel'} can be read as \textit{`I need a cheap hotel please, I am in a hurry'}. 
After that, we use the off-the-shelf ASR tool\footnote{https://www.alibabacloud.com/help/product/30413.htm} to transcribe all audios into noisy texts. Therefore, each dialog in RiSAWOZ is augmented into four dialogs with varied expressions and ASR noises. We carefully clean the texts to obtain our final SBD dataset. The comparison between SBD and  RiSAWOZ is shown in Table \ref{sbd_table}.

\begin{table}[t]
\centering
\scalebox{0.85}{\begin{tabular}{ccc}
\toprule
Dataset & \begin{tabular}[c]{@{}c@{}}RiSAWOZ\\ \cite{quan-etal-2020-risawoz}\end{tabular} & CGoDial-SBD \\
\midrule
\#Domain & 12 & 12 \\
\#Dialog & 11,600 & 44,800 \\
\#Turns & 75,991 & 473,348 \\
Extra knowledge & No & Yes   \\
OOS detection & No & Yes  \\
Spoken feature & No & Yes  \\
\bottomrule
\end{tabular}}
\caption{The comparison of between SBD and the RiSAWOZ dataset.}
\label{sbd_table}
\end{table}

\begin{figure*}[t]
    \centering
    \includegraphics[width=1.0\textwidth]{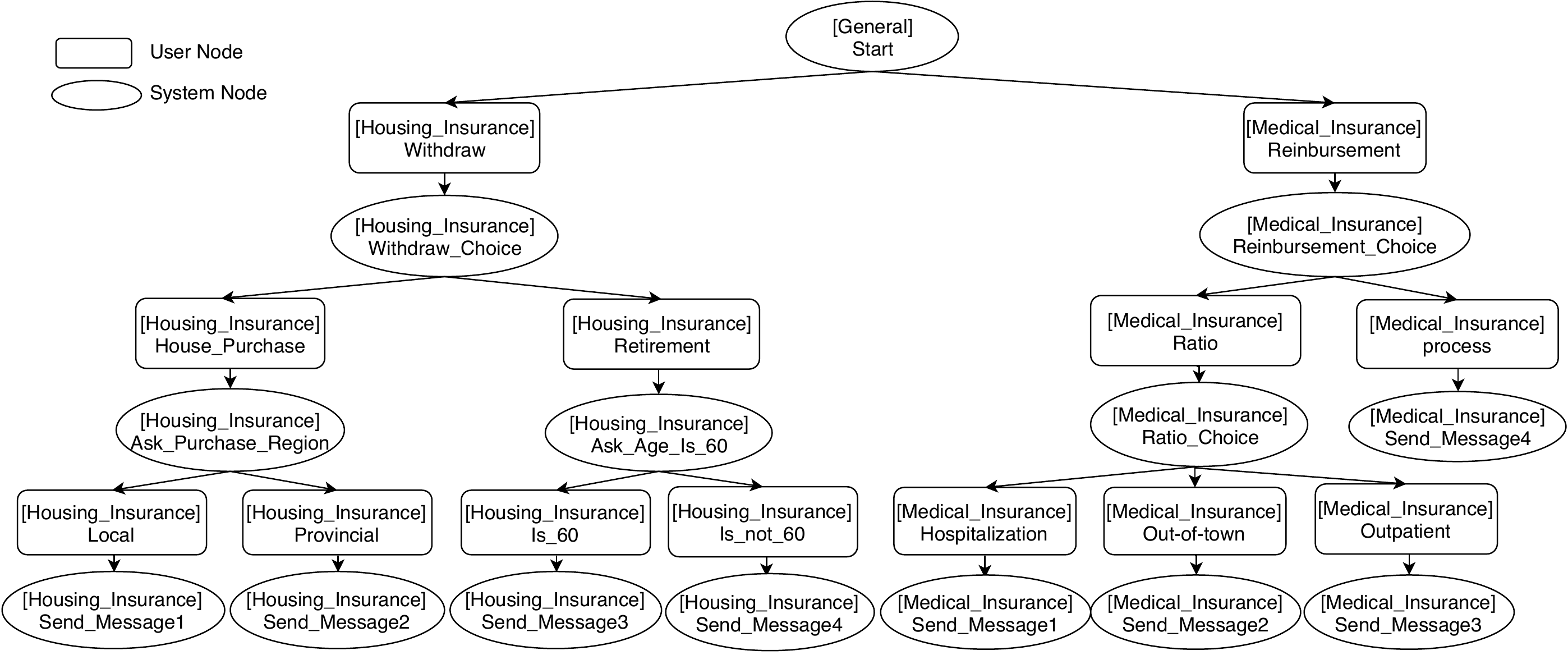}
    \caption{A part of the multi-domain FBD dialog flow from the  social insurance business (There are 284 nodes in total, we only show 27 nodes here). The names on the nodes are direct translations from Chinese. The words in the square brackets [..] are the \textit{domain} names, while the other words are the node names.
    Each user node can be viewed as a user intention. Each system node contains a predefined response.}
    \label{fig:flow_case}
\end{figure*}

\subsection{Flow-based Dialog Dataset}
Flow-based dialog (FBD) is quite common in industrial dialog products, such as such as Microsoft BotFramework \footnote{https://learn.microsoft.com/en-us/composer/}, Google DialogFlow \footnote{https://cloud.google.com/dialogflow}, Salesforce Converse \footnote{https://github.com/salesforce/Converse} and Alibaba Intelligent Robot \footnote{https://www.alibabacloud.com/product/bot}, due to its easy-to-use and drag-and-drop characteristics for dialog developers. However, currently, there is no available Chinese dataset for FBD, which motivates us to collect a new FBD dataset to facilitate the research. 

\subsubsection{Task Definition}
The flow-based dialog (FBD) typically contains an explicit grounded dialog flow to instruct the ongoing conversation between users and systems. The flow is a decision tree with user nodes and system nodes alternately. The user node (a.k.a., intention node) specifies the classes that what kind of utterance the user would say, the system node (a.k.a., reply node) specifies the system response. Each user node points to only one system node, but each system node can point to multiple user nodes. Usually, the flow is handcrafted by experienced dialog composers and specifies the strict order of the dialog process \cite{mosig2020star}. 
For example, in a banking transaction business \cite{chen-etal-2021-action}, if a user wants to withdraw the deposit, a bank clerk should first determine the user's identity by confirming his residency, then it would go into detailed business. Although the dialog process in FBD is not as flexible as in SBD, it has more advantages in specifying certain rules in the conversation. 

For FBD, the main task is to predict the correct user node on the flow that the current user utterance is talking about. Therefore it can be formulated as a node classification problem given the dialog context and the flow structure. All the responses on the system nodes are pre-defined, so both dialog policy and dialog generation tasks are not crucial. Inspired by \citet{rajendran-etal-2019-learning}, two essential metrics are used here: 
1) \texttt{TurnACC} measures the percentage of user nodes that are correctly chosen,
2) \texttt{DialACC} measures the percentage of dialogs where every user node is correctly chosen. The former can be viewed as an ability of dialog understanding, while the latter tracks task completion.

\subsubsection{Dataset Construction}
We first manually design four different dialog flows for daily businesses, including social insurance,  residency management, housing insurance, and electronic toll collection. Please refer to the Figure \ref{fig:flow_case} for a detailed dialog flow from social insurance. 
We then deploy four different dialog agents for the businesses using the Alibaba  Intelligent Robot. All the agents are then used to collect human-to-machine spoken dialogs from real conversations. After collecting the online conversational logs, we use the same ASR tool to transcribe the spoken audios into noisy texts to construct the dataset. We then ask two business experts to examine the dialogs and rectify mispredicted dialog data.

Detailed statistics of FBD are given in the Table \ref{tab: fbd_number}. As shown in the table, we propose a new FBD dataset with 6.78k turns spanning four different businesses based on specific business trees. The system will guide the user to finish the goal according to the schema described in the tree. The dialog progresses via traversing the tree, where at each turn, the agent needs to select the correct user node from all possible successors by understanding the user's utterance, and give the response saved on the next corresponding system node. 

\begin{table*}[tbp]
\scalebox{0.85}
{
\begin{tabular}{ccccccc}
\toprule
\multirow{2}{*}{Dataset} & \multirow{2}{*}{\begin{tabular}[c]{@{}c@{}}STAR\\ \cite{mosig2020star}\end{tabular}} & \multicolumn{5}{c}{CGodial-FBD} \\ \cline{3-7} 
 &  & \begin{tabular}[c]{@{}c@{}}Social \\ Insurance\end{tabular} &
 \begin{tabular}[c]{@{}c@{}}Residency \\ Management\end{tabular} & \begin{tabular}[c]{@{}c@{}}Housing \\ Insurance\end{tabular} & \begin{tabular}[c]{@{}c@{}}Electronic \\ Toll Collection\end{tabular} & \textit{Total} \\
 \midrule
\#Domain & 13 & 6 & 10 & 9 & 12 & {37} \\
\#Dialog & 5,820 & 4,510 & 1,036 & 454 & 786 & {6,786} \\
\#Turn & 127,833 & 17,510 & 4,488 & 1,316 & 3,035 & {26,349} \\
\#Node & 297 & 284 & 260 & 146 & 98 & {788}  \\
Spoken feature & No & Yes  & Yes  & Yes  & Yes  & -- \\
\bottomrule
\end{tabular}}
\centering
\caption{The detailed comparison of between FBD and STAR dataset.}
\label{tab: fbd_number}
\end{table*}



\begin{table*}[t]
\resizebox{0.95\textwidth}{!}{
\begin{tabular}{ccccc|cccc|cccc}
\toprule
\multirow{2}{*}{Model} & \multicolumn{4}{c|}{Full-train \& Test-std}                                 & \multicolumn{4}{c|}{Few-train \& Test-std}                                  & \multicolumn{4}{c}{Full-train \& Test-hard}                                       \\ \cline{2-13} 
                       & \texttt{JGA}           & \texttt{Succ}          & \texttt{BLEU}          & \texttt{Comb}          & \texttt{JGA}           & \texttt{Succ}          & \texttt{BLEU}          & \texttt{Comb}         & \texttt{JGA}           & \texttt{Succ}          & \texttt{BLEU}          & \texttt{Comb}         \\ \hline
Chinese-T5                     & 49.4          & 48.3          & \textbf{26.2} & 75.1          & 2.16          & 5.42          & 8.27          & 12.1          & 27.43          & 26.17          & \textbf{22.49} & 49.3          \\
CDial-GPT                    & \textbf{53.7} & 49.6          & 24.4          & 76.1          & \textbf{5.58} & 5.49          & 8.10          & 13.6          & 33.15          & 28.00          & 19.28          & 49.9          \\
Our PCM                    & 51.6          & \textbf{52.6} & 25.6          & \textbf{77.6} & 5.17          & \textbf{5.62} & \textbf{8.89} & \textbf{14.3} & \textbf{33.49} & \textbf{32.83} & 21.63          & \textbf{54.8} \\

\bottomrule
\end{tabular}
}
\centering
\caption{Results on slot-based dialog (SBD) dataset.}
\label{tab:SBD}

\end{table*}
\begin{table*}[htp]
\resizebox{0.95\textwidth}{!}{
\begin{tabular}{ccc|cc|cc|cc}
\toprule
\multirow{2}{*}{Model} & \multicolumn{2}{c|}{Full-train \& Test-std}   & \multicolumn{2}{c|}{Few-train \& Test-std}    & \multicolumn{2}{c|}{Full-train \& Test-hard}      & \multicolumn{2}{c}{Few-train \& Test-hard} \\ \cline{2-9} 
                       & \texttt{TurnACC}        & \texttt{DialACC}        & \texttt{TurnACC}        & \texttt{DialACC}     & \texttt{TurnACC}        & \texttt{DialACC}        & \texttt{TurnACC}        & \texttt{DialACC}           \\ \hline
StructBERT             & 77.13          & 55.45          & 63.79          & 40.44          & 55.40          & 13.34          & 50.60             & 13.09             \\
Roberta-wwm                & 78.01          & 55.68          & 67.28          & 44.24          & 55.02          & 12.69          & 53.16             & 14.58             \\
Our PCM                   & \textbf{78.97} & \textbf{58.80} & \textbf{69.72} & \textbf{46.91} & \textbf{57.73} & \textbf{15.03} & \textbf{56.37}    & \textbf{16.88}    \\ \bottomrule
\end{tabular}
}
\centering
\caption{Results on flow-based dialog (FBD) dataset.}
\label{tab:FBD}
\end{table*}
\begin{table*}[h]
\resizebox{0.95\textwidth}{!}{
\begin{tabular}{ccc|cc|cc|cc}
\toprule
\multirow{2}{*}{Model} & \multicolumn{2}{c|}{Full-train \& Test-std}   & \multicolumn{2}{c|}{Few-train  \& Test-std}    & \multicolumn{2}{c|}{Full-train \& Test-hard}      & \multicolumn{2}{c}{Few-train \& Test-hard}\\ \cline{2-9} 
                       & \texttt{\ \ R@1\ \ }       & \texttt{\ \ MRR\ \ }        & \texttt{\ \ R@1\ \ }       & \texttt{\ \ MRR\ \ }       & \texttt{\ \ R@1\ \ }       & \texttt{\ \ MRR\ \ }    & \texttt{\ \ R@1\ \ }       & \texttt{\ \ MRR\ \ }      \\ \hline
StructBERT             & 35            & 54.82          & 15.6          & 33.69          & 9.60           & 37.16          & 8.48               & 26.62            \\
Roberta-wwm                & 35.3          & 56.16          & 17.9          & 39.30          & 18.08          & 43.56          & 10.85              & 32.27            \\
Our PCM                    & \textbf{39.1} & \textbf{59.54} & \textbf{27.7} & \textbf{48.02} & \textbf{23.50} & \textbf{48.60} & \textbf{17.80}     & \textbf{39.7}    \\
\bottomrule
\end{tabular}
}
\centering
\caption{Results on retrieval-based dialog (RBD) dataset.}
\label{tab:RBD}
\end{table*}

\subsection{Retrieval-based Dialog Dataset}
\subsubsection{Task Definition}
Learning end-to-end retrieval-based dialog is a crucial  direction in dialog research \cite{bordes2016learning, dai-etal-2020-learning, tao2021survey}, which is also very common in real applications \cite{williams-etal-2017-hybrid,henderson-etal-2019-training}. 
The retrieval-based dialog (RBD) in our benchmark is similar to the well-studied bAbI-dialog \cite{bordes2016learning}. Given the dialog history, RBD aims to select the correct response from a candidate set of responses, learning goal-oriented dialog in an end-to-end manner.  Since RBD is a retrieval problem, we leverage the standard IR metrics  to evaluate model performance, including recall@1 (\texttt{R@1}) and mean reciprocal rank (\texttt{MRR}).

\subsubsection{Dataset Construction}
We adapt from the existing text-in-text-out E-commerce Dialogue Corpus \cite{zhang-etal-2018-modeling} to construct our RBD dataset.
Since the original corpus is too large to be incorporated as a part of our dialog benchmark, we run five BERT baselines with different random seeds to sort the data according to the average predict accuracy and choose a proportion (100k turns) of difficult dialog examples as our initial RBD. Then we ask the crowd-sourcing people to vote whether the dialog turn has ambiguous response selection; if not, we ask them to read out all the utterances to add spoken features through the same procedure in SBD construction to acquire the dataset. 


\subsection{Data Quality Control}
Crowd-sourcing brings noisy data and annotations. To guarantee the quality of collected dialog, for all datasets (i.e., SBD, FBD and RBD) in CGoDial, we ask another three crowd-sourcing people to vote whether the data is recognizable and its annotation is nonambiguous. Then we recollect the dialog samples that at least two people vote against (around 20k turns) until they meet our standard.

\section{Experiments}


\subsection{Baselines}
Pre-trained conversation models gain increasing research interest in dialog communities \cite{zhang-etal-2020-dialogpt, wu-etal-2020-tod, space1, space2, space3}. 
Our CGoDial benchmark targets evaluating the Chinese pre-trained conversation models (PCMs).  However, the number of published Chinese PCMs is quite limited. For SBD, we use the Chinese-T5 \cite{zhao-etal-2019-uer} and CDial-GPT \cite{wang2020large} as base models, and use MinTL \cite{lin-etal-2020-mintl}, a T5-based dialog model, and UBAR \cite{yang2020ubar}, a GPT-based dialog model, as the downstream dialog models respectively. Note that both MinTL and UBAR are specifically designed for datasets like MultiWOZ, which is similar to our SBD. For FBD and RBD, since all tasks are classification tasks, we choose two Chinese BERT-like pre-trained language models, StructBERT \cite{wang2019structbert} and Roberta-wwm \cite{cui2019pre} as baselines. StructBERT incorporates language structures into pre-training for deep language understanding. Roberta-wwm uses the whole word masking for MLM training.

\subsection{Methods}
In this work, we propose a new Chinese PCM by pre-training a UniLM on a sizeable open-domain Chinese dialog corpus for ten epochs. The corpus comprises nearly 100 million conversations collected from online textual dialog forums. Since the total number is vast for the textual dialogs, we apply the artificial ASR augmentation \cite{ma-etal-2020-charbert} rather than crowd-sourcing to add spoken errors.
In this way, our PCM can achieve better robustness than other pre-trained models only trained on plain texts.

For SBD data, all models take the dialog history as input and output the dialog state, the dialog action, and the response for evaluation as in  \citet{yang2020ubar}. For FBD data, all models take the dialog history as input and give the node prediction by adding a linear layer to the pooled $\mathtt{[CLS]}$  representation for classification. We leave the utilization of tree structure in future work. For the RBD data, we concatenate the dialog history and each candidate response iteratively to predict a binary label $1/0$ by adding a linear layer to the pooled $\mathtt{[CLS]}$ representation, indicating whether the candidate is true or false. In the testing phase, we rank the candidates according to their predicted probability.

\subsection{Settings}
To measure how well the model performs on complicated dialogs, we pick hard test data from the three datasets in CGoDial to form a specific subset for robustness evaluation. We first run the three baselines (Chinese-T5, Roberta-wwm, and our PCM) to make predictions on all test data, and choose the dialog data with at least one baseline that gives the wrong answer. Then we ask crowd-sourcing people to score the data ranging from 1 to 5 according to several criteria: 1) whether it is hard to understand; 2) whether it contains spoken features (e.g., ASR errors) on keywords in utterances. Finally, we choose data scored more than 2.5 as our hard test set. 

The sizes of all the pre-trained models are the base scale (12 layers and
the 768 hidden embedding dimensions). For SBD, we use the same hyperparameters in the MinTL model for Chinsese-T5 , and the UBAR model for CDial-GPT and our PCM. For FBD and SBD, we use AdamW optimizer for optimization with an initial learning rate of 1e-5. The warm-up proportion is 10\%. The batch size is set to 128 and the maximum input length is 512. The hidden size of the output classification head is 128. Best model checkpoints are chosen based on the validation within 20 epochs.

\subsection{Results}
To evaluate the ability of  few-shot adaptation \cite{geng-etal-2019-induction, geng-etal-2020-dynamic} and robustness \cite{peng-etal-2021-raddle},  for all datasets in CGoDial, we employ several evaluation settings: 1) \textit{full-train\&test-std}, i.e., using all training data and evaluating on the standard test set. 2) \textit{few-train\&test-std}, i.e., using only 10\% training set and evaluating on the standard test set. 3) \textit{full-train\&test-hard}, i.e., using \textit{full-train} and evaluating on the hard test subset. 4) \textit{Few-train\&test-hard}, i.e., using \textit{few-train} and evaluating on the hard test subset.

Table \ref{tab:SBD} shows the results on slot-based dialog dataset. As we can see, our PCM achieves the best combined scores on different settings, especially in the test-hard setting. 
Concretely, our model outperforms other baselines on almost all the \texttt{JGA} and \texttt{Succ} metrics, indicating that it has good ability in dialog understanding and policy. Especially in the robust setting, our model obtains 4.83 points improvement (28.00$\xrightarrow{}$32.83) in \texttt{Succ} and 4.9 points improvement (49.9$\xrightarrow{}$54.8)  when evaluating on the test-hard set. We conjecture that our large-scale and harder training corpus makes our PCM perform more reliably in difficult spoken cases. 
However, our model performs slightly worse on the \texttt{BLEU} metric than Chinese-T5, possibly because UniLM is more suitable for language understanding tasks. When training in the few-shot setting, all models degrades drastically, showing a large space to improve the model few-shot learning ability in spoken task-oriented dialog tasks.
Since the results of \textit{few-shot\&test-hard} are too low, we neglect this hybrid setting.

Table \ref{tab:FBD} and \ref{tab:RBD} show the results on flow-based dialog and retrieval-based dialog datasets, respectively. In all settings, our PCM shows its superior performance to a large margin, demonstrating the ability of our PCM on classification dialog tasks in CGoDial. Particularly, on the  \textit{few-shot\&test-hard} setting, our PCM obtains  2.3 points improvement (14.59$\xrightarrow{}$16.88) in \texttt{DialACC} on FBD, and 6.95 points improvement (10.85$\xrightarrow{}$17.80) in \texttt{R@1} on RBD.
However, all tasks are still far from an acceptable performance for real applications, which indicates the challenge and difficulty of our benchmark.

In addition, to show that our augmented CGoDial datasets do have more difficulty in language variation and ASR error than non-augmented original dialog data, we run PCM on the \textit{full-train\&test-std} setting for each dataset and demonstrate the difference in Table \ref{tab: augment}.

\begin{table}[t]
    \centering
    \scalebox{0.75}{
    \begin{tabular}{c|c|c|c}
        \toprule
       Dataset  & metric & after augmented & before augmented \\
       \midrule
       SBD  &\texttt{Comb} & 77.64  &   90.32\\
       FBD & \texttt{DialACC} & 58.8 & -- \\
       RBD & \texttt{R@1} & 39.14 & 48.59  \\
       \bottomrule
    \end{tabular}}
    \caption{Comparsion of CGoDial and original data. FBD does not have comparison since it is collected from real spoken dialogs originally.}
    \label{tab: augment}
\end{table}

\section{Conclusion}

We propose CGoDial, a new large-scale Chinese goal-oriented dialog benchmark featuring three different types of datasets: slot-based dialog, flow-based dialog, and retrieval-based dialog, which are used for comprehensive dialog evaluation.
We also propose three different experimental settings: standard training, limited data use, and robustness testing, to assess the dialog model from the aspects of general prediction, fast adaptability, and reliable robustness, respectively. Results from several competitive baselines show the challenge of CGoDial, which is worthy of follow-up research.

\section*{Ethical Considerations}
The collection of our CGoDial dataset is consistent with the terms of use of any sources and the original authors’ intellectual property and privacy rights.
The new dataset (FBD) and adapted dataset (SBD, RBD) are collected with the ALIDUTY platform\footnote{Unfortunately,  the platform is closed now.}, and each crowd-sourcing person requires up to 10 minutes to complete. The source of annotators are mainly from colledge students and professional annotators provided by the platform.   
The requested inputs are general language variations and speaking voices. No privacy-related information is collected during data collection. Each person was paid 0.1-0.2 USD for a single turn dialog data, which is higher than the minimum wage requirements in our area. The platform also hires professional reviewers to review the collected data to ensure no ethical concerns, e.g., toxic language and hate speech.

\section*{Limitations}
From the aspect of the dataset construction, our CGoDial has two main limitations: 1) lacking human-to-human spoken dialogs in real situations. This kind of dialog data has many issues in terms of privacy; thus, it is not easy to open source. 2) For SBD and RBD, we re-build the dataset on other existing textual corpora; therefore, the original datasets largely restrict the final dialog-level flexibility and pattern variation.

From the aspect of the proposed method, we have not used tree structure in the FBD; however, the structural information should be crucial to improve the final overall performance.

\bibliography{anthology,custom}
\bibliographystyle{acl_natbib}

\clearpage

\appendix

\section{Appendix A.1}
\noindent
The dataset distributions of dialog length and turn length are shown in the following figures.

\begin{figure*}[hp]
    \includegraphics[width=1\textwidth,height=0.5\textwidth]{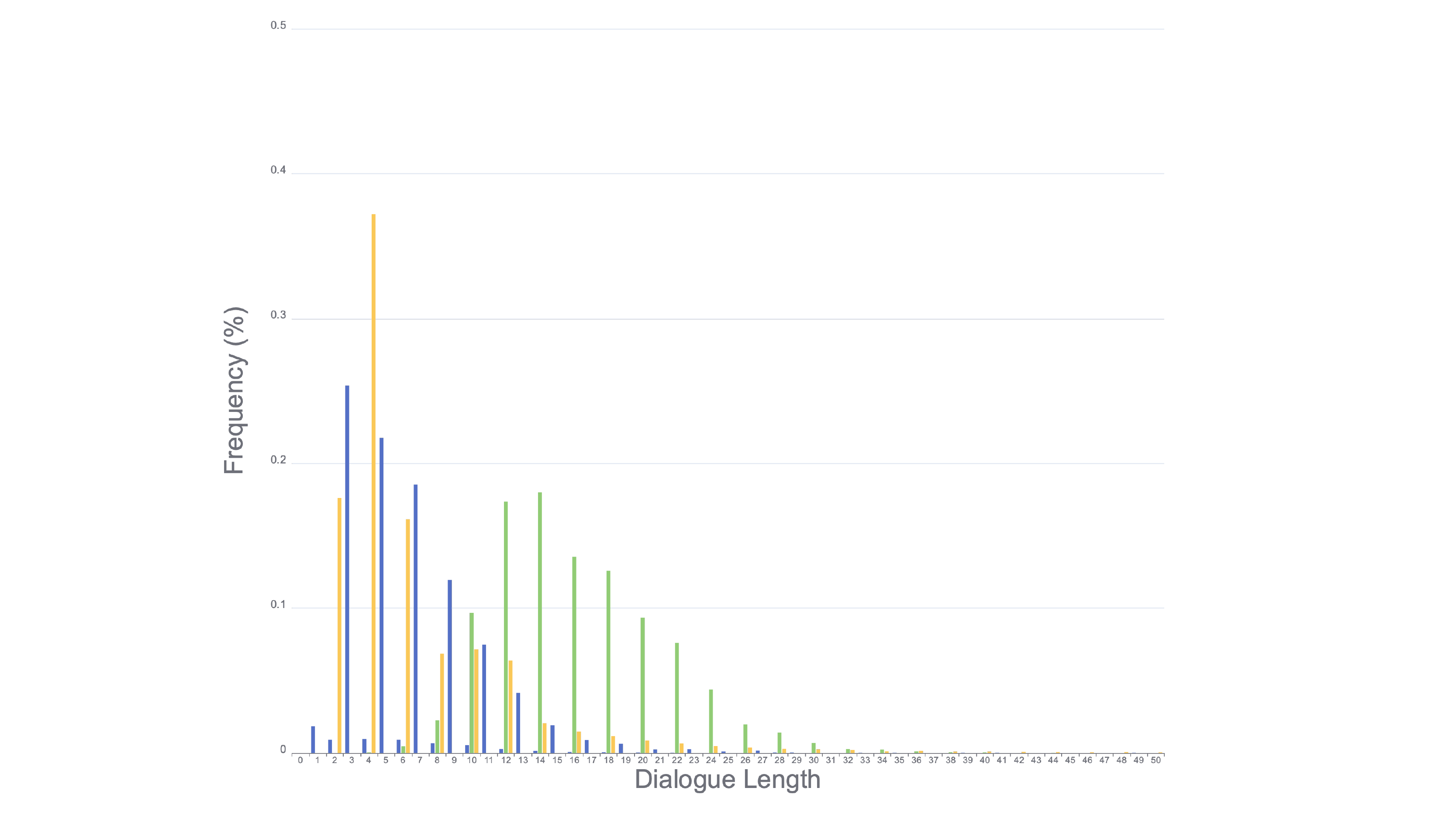}
    \caption{The distribution of the number of turns in the three kinds of dialog in CGoDial: \textcolor{BlueViolet}{flow-based}, \textcolor{YellowGreen}{slot-based} and \textcolor{Peach}{retrieval-based dialog}.}
    \label{fig:datasetdist}
\end{figure*}
\begin{figure*}[h]
    \centering
    \includegraphics[width=1\textwidth,height=0.5\textwidth]{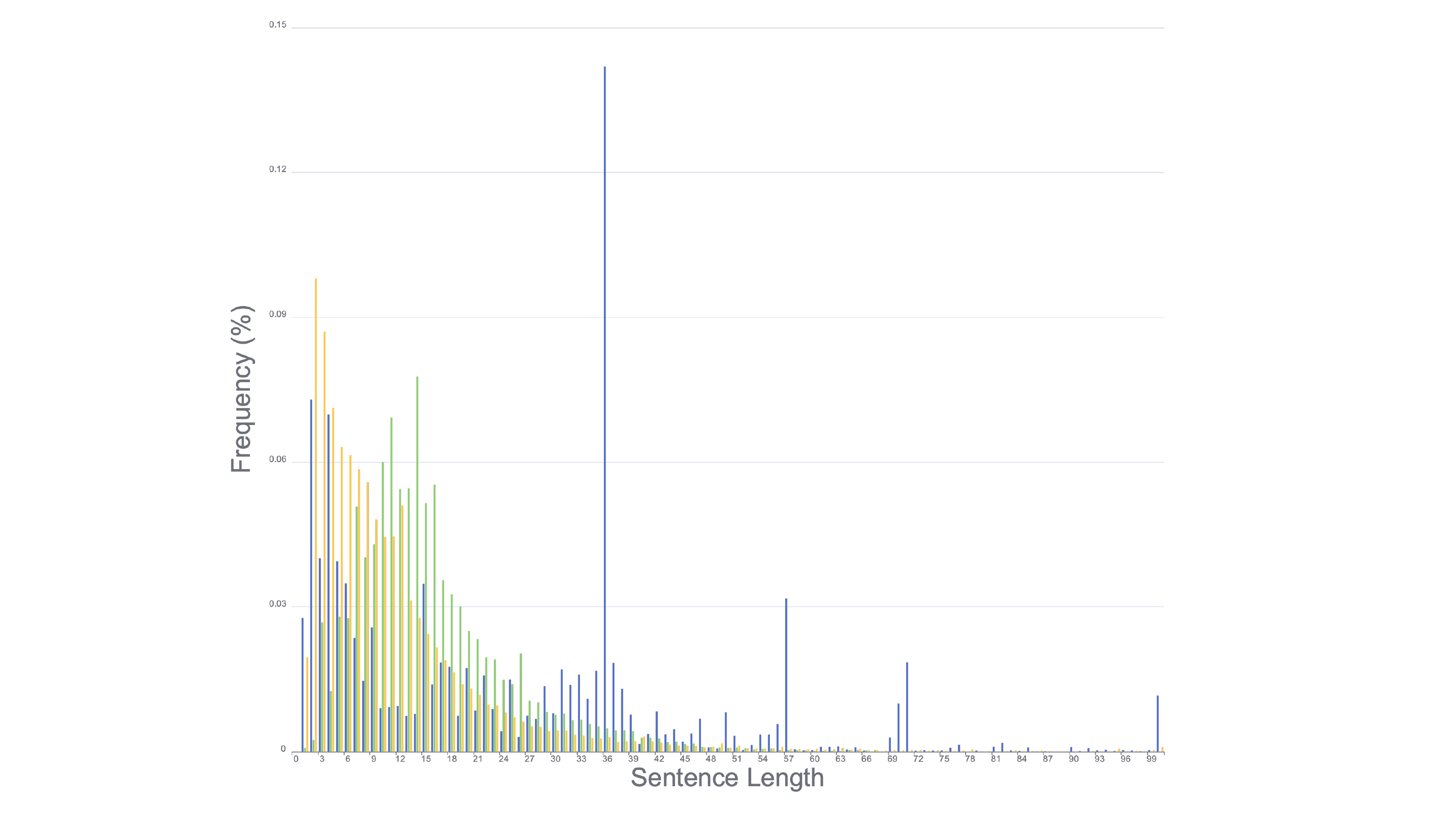}
    \caption{The distribution of the length of turn in the three kinds of dialog in CGoDial: \textcolor{BlueViolet}{flow-based}, \textcolor{YellowGreen}{slot-based} and \textcolor{Peach}{retrieval-based dialog}. The reason for some peaks in flow-based dialog distribution is due to the template response.}
    \label{fig:datasetdist}
\end{figure*}

\end{document}